\documentclass[a4paper]{book}

\usepackage{array}
\usepackage{amsmath}
\usepackage{graphicx}
\usepackage{thumbpdf}
\usepackage[pdftex,
        colorlinks=true,
        urlcolor=rltblue,       
        filecolor=rltgreen,     
        linkcolor=rltred,       
        pdftitle={Genetic algorithm for robotic telescope scheduling},
        pdfauthor={Petr Kub\'{a}nek},
        pdfsubject={Scheduling using Genetic Algorihms},
        pdfkeywords={autonomous observatory, robotic telescope},
        pdfproducer={pdfLaTeX},
        pagebackref,
        pdfpagemode=None,
        bookmarksopen=true]{hyperref}
\usepackage{color}
\definecolor{rltred}{rgb}{0.75,0,0}
\definecolor{rltgreen}{rgb}{0,0.5,0}
\definecolor{rltblue}{rgb}{0,0,0.75}

\usepackage{float}

\newcommand{\degree}{\ensuremath{^\circ}}

\begin{document}

\frontmatter

\pagestyle{empty}

\begin{center}

\vskip 9cm

\includegraphics[scale=0.35]{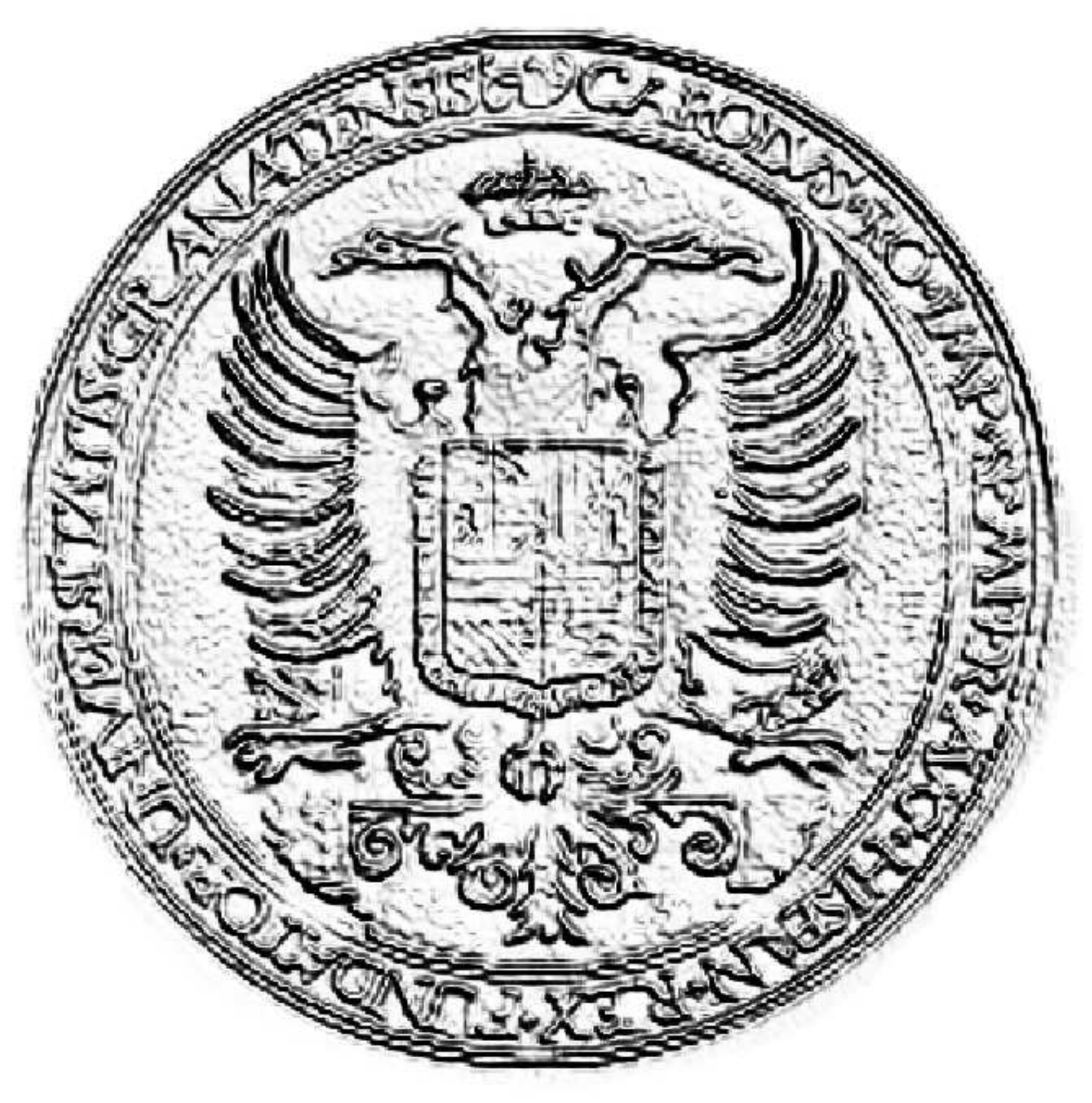}

\vskip 1cm

{\Large {\bf \LARGE Universidad de Granada}

\vskip 0.5cm

M\'aster en Soft Computing y Sistemas Inteligentes

}

\vskip 2cm

Trabajo de Investigaci\'on Tutelada

\vskip 0.5cm

{\bf \LARGE Genetic Algorithm for Robotic Telescope Scheduling}

\vskip 0.5cm

Curso acad\'emico MMVII / MMVIII

\vskip 4cm

\begin{tabular}{m{0.5\linewidth}m{0.5\linewidth}}
{\begin{centering} Profesor \\
Dr. D. Francisco Herrera \\
Dr. D. Alberto J. Castro-Tirado \\
\end{centering}
}
&
{\begin{centering} Realizado por \\
Petr Kub\'anek \\
\end{centering}}

\end{tabular}

\end{center}

\newpage

\mbox{}

\paragraph{Acknowledgement}

I would like to acknowledge financial support from Spanish Ministry of
Education and Science via grant BES-2006-11506. This work benefits from my
experiences gained developing RTS2. RTS2 development was encouraged and
supported by many persons, among them worth noting are Alberto Castro-Tirado,
Martin Jel\'inek, Ronan Cunniffe, Ren\'e Hudec and Victor Reglero. This work
will never be produced without kindly help and assistance of professors
Francisco Herrera and Antonio Gonz\'ales. And I am gratefull to Rocio Romero
for providing me \mbox{NSGA II} description.

\tableofcontents

\newpage


\pagestyle{headings}

\chapter{Preface}

This work was inspired by author experiences with a telescope scheduling.
Author long time goal is to develop and further extend software for an
autonomous observatory. The software shall provide users with all the
facilities they need to take scientific images of the night sky, cooperate with
other autonomous observatories, and possibly more. It shall provide support for
as many devices coming from as many different vendors as possible, yet remains
"plug and play" to setup and operate.

From thirty thousands feet view, telescope time scheduling looks simple. Take
some entries, select the best one, and observe it as long as possible. If it is
no more worth to observe this particular position, choose another. Do so every
night from dusk to dawn, see results of the night runs, analyse acquired images
and try to advance mankind knowledge of the astronomy, astrophysics or physics
from the acquired images.

However, as we zoom closer to provide more details about scheduling, nice flat
landscape become covered with various obstacles. Start with the definition what
is the best target -- is it the one which is currently rising or the one which
is setting and so telescope will not have possibility to observe it again for
some time. And what about future plans? Is it better to observe star A, and in
a few weeks time star B, or shall the observatory pick star A and B at the same
night?

No wonder that scheduling is one of the major issues of current world class
observatories, both ground and space. Usual approach is to have some time
allocation committee, which pick which observations proposals are worthy to
observe, and then meets every day, week, month or year to discuss which objects
will be observed during the next period. Rules are created for sharing risks
between an observatory and the observation proposal authors -- for example it
has to be defined what shall happen if bad weather prevents observations on a
ground observatory, or if data are lost during transmission from a satellite to
the ground. And humans, which are member of time allocation committee, make
sure that all observation constraints are obeyed and that users are happy with
acquired data and do a lot of interesting and useful science based on the data
acquired.

Autonomous observatories are by definition autonomous. They do not have any
time allocation committee composed of smart people. They are destined to create
observation strategy from inputs which human operators provided to them and/or
from results of the previous observations. This works provides overview of the
various constraints and objective of the observatory, and gives some solution
to the scheduling problem.

The obvious question which can be asked is why we need the schedule at all?
Would not it be better to simply pick the best observation, and then the next
one, as mentioned above? Is not this a principle of the queue scheduling, which
is used on most of the human operated observatories today? There are various
hints which gives us answer that this is not a desired operation for the
autonomous operations.

The first one comes from possible network scheduling. Network nodes can of
course by scheduled locally. But that will not exploit network advantages.
Network can observe one star simultaneously over period longer then usual dark
period at a single site. For this kind of observations some collaboration among
schedulers is needed. One of the possibility is to let local scheduler
collaborate and create a plan. Second possibility is to have possibility to
centrally schedule network observations. Our experience shows that while first
option is feasible, second is more easily controlled and monitored by a human
operator.

The second case come from operation of a single instrument. Our experience
shows that it is handy to have a schedule before start of the night to check
what observatory will do during night. The system poses option to create an
observing plan and then to execute it, but plan has to be created somehow. And
as the scheduling process is annoying and can only distract person doing it
from other work, plan creation has to be automated. With human experts possibly
reviewing it and taking actions if they found something strange on the
schedule.

\mainmatter

\chapter{Introduction}

RTS2\cite{2006SPIE.6274E..59K} is an open-source software package for robotic
observatory. Apart from a central server, device drivers and various other
functions, it provides service for target selection - a selector.

Current selector uses a simple single merit function selection to select next
observation from the list of the possible observations. The merit function,
which measure how good is a target for observation, is hard-coded in selector
code and cannot be easily modified.  Moreover, current selector does not allow
generation of an observing plan for a full night.  Current design allows only
selection of the next target. After the current target observation is finished,
the target with the highest current merit function value is selected by
the selector. Creating night schedule for telescope belongs to NP class of
problems. Some heuristic might apply, but the problem will remain hard to
solve.

This work describes design and implementation of a selector based on genetic
algorithm. This selector allows generation of the full night plan. It must also
keep an overview of which important targets remain to be observed, and allow to
schedule them appropriately. The algorithm must be extensible for scheduling of
the robotics telescope network. And it must be written so it can be easily
integrated with the current \mbox{RTS2} code.

\section{Basic definitions}

{\bf Target} is an object on the sky telescope users would like to observe. It
can have fixed coordinates (stars,..) or its coordinates can change with time
(planets, solar system minor bodies,..). For given time and location, target
altitude, zenith distance and azimuth can be easily calculated.
 
{\bf Observation} is one visit to the target. During observation, target can be
observed in different colours, using filters placed in optical path by some
rotating mechanism. If an observation does not satisfy user needs (e.g. does
point to clouds, instrument problems occured during observing run,..) it shall
be rescheduled.  Observation usually consists of images acquired with different
filters. For some targets the length of the observation can be modified, for
others it must remain fixed.

\section{Constrains}

The following constrain might apply to a target:

\begin{itemize}

\item target must be {\bf visible}, e.g. above local horizon

\item various constraints based on {\bf lunar distance} of the target, solar
distance, moon phase (some targets cannot be observed with a full moon, etc..).

\item {\bf budget constraints} -- time allocated to target/user can be limited,
and target cannot exceed it

\item {\bf time} -- some targets required observation in a given time, some can
be observed anytime they are visible, some targets should be observed only once
per night, week, .., as other observations will not lead to increase in the
scientific return obtained from the acquired images

\item when scheduling a network of autonomous observatories, some observations
can be carried either simultaneously with two or more telescopes, or
observation with one telescope must follow observation with another telescope

\end{itemize}

Constraints are formalised in the third chapter.

\section{Selection of the best schedule}

Astronomers would measure an observatory success by a number of the articles
which reference the observatory (possibly weighted by the journal citation
index). Because scheduling does not pose black magic it is unable to predict
which targets observations will contribute to the most interesting discoveries
and scientific papers. So some evaluation function has to be defined which
compares targets. The goal would be to found a sequence with the highest sum of
observation fitness values, while keeping the constrains outlined above.
Evaluation can be based on:

\begin{itemize}

\item {\bf height} of the target during observation (higher is usually better)

\item {\bf distance} of the target from the Moon or any other sky body

\item {\bf quality} of the proposal (some accepted targets may only fill the
gap, other can be ranked as top priority by evaluation committee)

\end{itemize}

Formalisation of the various fitness measurements is described in the third
chapter.

\section{Genetic algorithms for robotic telescope scheduling}

Aim of this work is to design, develop and tests various genetic algorithm for
scheduling of the robotic observatories. Problem is similar to job shop
scheduling, and thus belongs to NP-hard problems. Schedule for one telescope
and scheduling of telescope network will be investigated.

Problem input consists of targets and their characteristic. Target is
location on the sky which can be visited by telescope to acquire data.
Each target have constrain which specify when it can be observed.

Output is list of targets and their observations times on the telescope
or on the telescope network.

\chapter{Autonomous robotic observatory}

Autonomous robotic observatory is a complex environment of computers, networks
and instruments. Aim of an autonomous robotic observatory is to carry
observations of night sky and record the observations using CCD cameras or
other instruments.

Most of the modern astronomical observatories are controlled by computer. But
degree at which computer is in charge of the observatory varies significantly.

One extreme is a computer controlling only the telescope, and offering observer
quick way how to enter target coordinates. Observer is then responsible for
scheduling, acting as coordinator synchronising various instruments, recording
data and processing them.

Other extreme end of observatory a complete autonomous observatory.  Software
is then responsible for observatory operations, opening and closing of the
roof, assuring that observatory is protected from elements, and reducing data.

Only a few world observatories are operated in a fully autonomous mode, without
night operator overseeing night operations. Full list of observatories which
claims to be robotic is provided at \cite{Hessman}. It should be noted, that
\mbox{RTS2} operates at least 6 telescopes with others coming hopefully in near
future. And at least 2 of those were run in fully autonomous mode over periods
of years - FRAM and Watcher.

Major current projects which operates and further extend fully autonomous
observatories \footnote{of which author of this work is informed} are "{\it
Thinking Telescope/RAPTOR}" developed at LANL\cite{ttlanl}, AudeLA developed at
Observatory de Haute Provance\cite{audela} and Las Cumbres Observatory Global
Telescope Network\cite{lcogt} -- eStar project\cite{estar}, which operates
Faulkes Telescopes. It should be noted, that from those only eStar is actively
investigating telescope scheduling\cite{fraser}, and AudeLA is the only other
project which provides observatory control system under open source license.

Majority of the world observatories operates in semi automatic mode. Scheduling
is usually done by human in the loop, supported by tools to help him/her decide
the best strategy. Following paragraphs provide review of the current practice
at leading Spanish and European observatories.

\section{CAHA}

Calar Alto Hispano Aleman (CAHA) observatory currently operates two major
instruments - 3.5m and 2.2m telescope. Both are remotely controlled by night
operators. For troubleshooting, operators are equipped with a torch, two way
radio and a car to drive to the instrument. Some scripting is provided for some
observations, but it is a night assistant who is responsible for operating the
instrument.

Scheduling on 2.2m is done on paper basis, with observer having printed
observing proposal for a night and selecting two or three observations he will
be performing during night. He is then responsibly to enter targets to the
schedule and oversee that observations are performed as expected.

Scheduling of 3.5m is even more complicated. Night staff have printed observing
proposals, pick up the one that will be observed and according to proposal text
handle instrument setup and observation synchronisation.

\section{OSN}

Observatory de Sierra Nevada has three major instruments -- 1.5m, 0.9m and 0.6m
telescope. The 0.6m is controlled by RTS2, so it is designed to be fully
autonomous. The other telescopes are controlled by night observer, who either
carry all observations himself, picking targets from a prepared list, or enter
current target to observatory control system and check that the observations
are performed as expected.

\section{ESO VLT}

European South Observatory Very Large Telescopes are operated in queue
scheduling\cite{vlt}. Night observers have screen with preselected list of
possible observations. Depending on observing conditions, their experience and
mood they select and oversee progress of the observation they choose to
perform.

\section{GTC}

Grand Telescopio de Canarias is now in commissioning phase. So far queue
scheduling is envisioned once telescope will be open for scientific
observations. Software for telescope operation posses similarities with ESO
BOSS\cite{vlt}, mentioned above.

\chapter{Formalisation of the observation scheduling problem}

The problem deals with distributing time on a single instrument. Night is time
when observatory is operational and can take observations. Schedule is sequence
of targets which will be observed during night. Each target have position where
it will be observed, observation script which defines how the observations will
be observed - for example which filters and exposures combinations will be used
during observation. For each target, various properties are set, and other
properties can be calculated.

In the following paragraphs are defined terms that will be used when dealing
with scheduling problem.

\section{Night}

Night start at some time after sunset and ends at some time before sunrise.
More complicated scheduling scenarios might include use of twilight period,
when observation is possible on certain parts of the sky. To keep problem
simply it is assumed that night runs from time $N_s$ till time $N_e$. Night
has then duration $N_d = N_e - N_s$.

Observatory operation can be disturbed by various factors. Those can be divided
into predictable and unpredictable interruptions. Predictable interruptions are
usually caused by maintenance work, which must be performed at given time at
the observatory. Unpredictable is weather, which causes major observatory
downtime, and technical issues with the observatory, which causes some
downtime. Depending on various factors (location, season, ..) weather usually
account for downtime between few percents up to 100\%. But observatories are
usually not build on sites where back weather account for more then 70\% of
available night time. Technical downtime is on tuned--up systems less then 1\%.

Unless explicitly specified, a case of an ideal observatory without any
downtime is considered.

\section{Observing sequence}

Observing sequence describes how the observation of a target is carried.  It is
a sequence of camera exposures, telescope operations and various modifications
performed on instruments on light path. For some targets, observing sequence
can be looped. For others, only single observation must be carried.

\section{Target}

Target is a position on sky which can be observed. Target has sky location,
usually expressed in equatorial coordinate system as right ascension and
declination. Each target has assigned observing sequence.

Observing sequence can change depending on various parameters. But that change
will make problem even more complex. Unless explicitly stated, only case of a
single observing sequence for a target is considered.

Targets are included in set $TS$. Size of target set is equal to $|TS|$.

\section{Observation}

Observation is a single visit of the telescope of a single target location.
Data are acquired during observation. Observing sequence describes how the data
shall be acquired.

\section{Duration of target observation}

Each target have three major duration values. When combined together they
describe how much night time will be used by a single target observation. Slew
time, $T_s$, describes how much time will be spend by slewing telescope on
target. Shutter open time, $T_o$, gives total time of the exposures taken for a
single observing sequence of target. Total observation time, $T_t$, gives time
spend in a single observing sequence. If telescopes moves during observing
sequence, $T_t$ contributes to total observing time and not to slew time -- slew
time is only the time needed to perform first slew to target.

Observing sequence can be repeated $l$ times, where $l \geq 1$. Dark time $T_d$,
time when shutter is closed, is equal to

\[
T_d = T_s + l * (T_t - T_o)
\]

Target slew time depends on previous telescope position. For some targets,
observing sequence can be looped and so total observing time can change in
multiplies of observing sequence duration. For others targets, only single
observing run must be performed, and so total observing time cannot change.

Total observation time $TT$ is then calculated as

\[
TT = T_s + l * T_t
\]

\section{Observation fitness}

Observation fitness describes how good is it to observe a target at a given
time. It is important to realize that the position of the target on the sky and
its distance to various disturbing bodies depends on time. Hence observation
fitness depends on time. So the observation fitness can be described as
function:

\[
f: time \Rightarrow fittness
\]

where time is time variable and fitness is some arbitrary set which describes
target fitness.

To make further explanation more understandable, we divide the fitness function
into two parts. The first depends only on time, second depends on a target
position. Following two paragraphs describes various fitness functions. The
first paragraph describes those which depends primarily on time, the second
those which depends on target position.

For algorithms to transform object coordinates to object position at a given
time, and to calculate object position with respect to other bodies, please
refer to \cite{aalgo}. Please see libnova (\cite{libnova}) for their
implementation.

\section{Observation time fitness}

To formalise observation time fitness binary logic is used. It is either
interesting or uninteresting to observe target at a given time. So the fitness
function is one returning either 0 or 1:

\[
f_t (t) = 0 \vee 1
\]

The $f_t$ function can depend on various factors. Even through the author of
this text gain some experience in the area, full list of those factors is
beyond his current knowledge. The ones he can mention are: {\bf time from
last observations of the object}, {\bf brightness of the object which show
periodic brightness variations} and {\bf special observing circumstances}.

Time dependent brightness variability of the objects observed by the
astronomers can be separated into three classes: regular time variability in
brightness, irregular time variability and brightness variability bellow
detection limit of the instrument. The objects can also show regular time
variability with superimposed irregular time variability.

Objects without any significant brightness variability have time fitness
constant through whole night. Objects which shows regular time variability are
usually worth observing at a certain time in the variability period.  Hence
those objects should have time fitness higher when it is worth observing them.

Objects showing irregular time variability can be observed anytime. However, if
instrument or other astronomers detects that the object of interest is showing
some interesting behaviour, usually increase in brightness, they shall be
visited more frequently.

\subsubsection{Time from the last observation of an object}

This case can be used when astronomer would like to monitor the object behaviour
in predefined intervals. If he/she is interested in variability of the object
at time scale of $t_{var}$ seconds, then the observations shall usually be
carried every $\frac{t_{var}}{2}$ seconds. So the $f_t()$ function
will be written in form:

\[
f_t (t) = 1 \iff now - T_s \geq \frac{t_{var}}{2}
\]

where $now$ is current time and $T_s$ is time of start of the last observation
of the target.

\subsubsection{Phase of the object which show periodic brightness variations}

Suppose that an object has periodicity $P_l$ seconds. Suppose that one know
period started at time $P_s$. Astronomer is interested in data taken in phase
between $H_s$ and $H_e$. Then time fitness function will then become:

\[
f_t (t) = 1 \iff \exists h \in Z : now - P_s - h * P_l \in < H_s, H_e >
\]

\subsubsection{Special observing circumstances}

Special observing circumstances are some know circumstances which will occurs
and which will make target observation interesting. For example, consider
transit of some solar system body in of some bright background stars. The
transiting body can be as big as the Moon and as small as some minor solar
system body. If astronomers have precise timing of transit, they can calculate
object size and others interesting parameters.

Of course not all targets shows this dependency. For them, the special time
fitness function $f_s$ is equal to 1. For those which have special time
dependence, there is set of times when observation should start. The duration
of observation is governed by observing sequence. To formalise this, we have a
set $TC$ of pairs ${T_s, T_e}$. Then fitness function $f_s$ is defined as:

\[
f_s (t) = 1 \iff \exists s \in TC : t \in s
\]

\section{Observation position fitness}

Following paragraphs describes some of the time fitness functions, which
depends on target position. As was mentioned in introduction to this chapter,
position of the target can be calculated from time and target properties.

First some introduction to how target position can depend on a time is given.
Then various factors which affect target fitness depending on its position are
described.

Objects observed by astronomers are located on the stellar sphere. As earth
rotates, those objects show apparent movement on the sky. Furthermore position
of objects which are close enough to the Earth and moves significantly in
respect to the Earth changes with regard to the stellar sphere. For example
objects in the solar system -- planets, dwarf planets and other solar system
bodies -- moves on the sky with comparison to more distant background stars.
Satellites on the Earth orbit moves even more quickly then the solar system
bodies.

\subsubsection{Relation between object position and Moon position}

Moon significantly increases sky brightness. Increased sky brightness can make
some observations useless, and other more difficult to process. The targets can
have following constraints:

\begin{itemize}

\item target cannot be observed if Moon height is above certain limit

\item target cannot be observed if Moon phase is in certain interval and Moon
height is above certain limit

\item those which can be observed only if their distance to the Moon is above
certain limit and Moon phase is above certain limit

\end{itemize}

As Moon sky position changes roughly by $13^\circ$ in 24 hours, object distance
to the Moon does not changes significantly during night. Moon phase, if
measured in range $<9^\circ, 360^\circ>$, changes by same amount. This show
that fitness function based on moon position, $f_m$, will not show great
variance during night if it depends only on distance of the object to the Moon.
So $f_m$ with only two possible values, $0$ or $1$, is used. So for a target
with duration $TT$, $f_m$ is defined as:

\[
f_m(t) =
\begin{cases}
1 & $if all moon constraints are valid in time $ <t, t + TT> \\
0 & $otherwise$
\end{cases}
\]

\subsubsection{Object altitude}

Object altitude changes during the night as it moves on the sky. For an object with
{\it declination} $\delta$ and hour angle $H$, and observing site with latitude 
latitude $\phi$, the altitude $h$ of the object is calculated as:

\[
sin(h) = sin(\phi) * sin(\delta) + cos (\phi) * cos (\delta) * cos (H)
\]

The minimal altitude of an object attained during 24 hours $Aday_{min}$ is
for an observer on northern hemisphere calculated as

\[
Aday_{min} = \phi - 90^\circ - \delta
\]

The maximal altitude $Aday_{max}$ is equal to 

\[
Aday_{max} = 90^\circ - \phi + \delta
\]

For an observer on southern hemisphere signs in the above formulas before
$\phi$ and $\delta$ has to be swapped.

As objects can be observed only during night, more important are values of
$Anigh_{min}$ and $Anigh_{max}$, the maximal and minimal altitudes of the
object during night. Those are calculated as

\[
Anight_{min} = 
\begin{cases} 
  Aday_{min}  & if \  t_{lower transit} \in < N_s , N_e > \\
  \min (A_{N_s}, A_{N_e})  & otherwise
\end{cases}
\]

\[
Anight_{max} = 
\begin{cases} 
  Aday_{max}  & if \  t_{upper transit} \in < N_s , N_e > \\
  \max (A_{N_s}, A_{N_e})  & otherwise
\end{cases}
\]

Those calculations reflect fact that minimal or maximal altitude is reached
either during given interval or on one of its edges.

Each target $T$ has low observing altitude $TA_{min}$. It is useless to observe
the object bellow this altitude. So if $h$ is target altitude, then
position fitness function is equal to 0 if $h \leq TA_{min}$.

If we do not consider changes in weather, which might render target observation
useless, then the best time for target observation is when its altitude is
maximal. To formalise this, height fitness function $f_h$ has range <0,1> and
is calculated as

\[
f_h(h) =
\begin{cases}
0 & if h \leq TA_{min} \\
\frac{h - max (Anight_{min}, TA_{min})}{Anight_{max} - max (Anight_{min}, TA_{min})} & otherwise
\end{cases}
\]

\section{Observation accounting}

Observatory time is usually shared by multiple groups. They contribute to
capital and operational costs of the observatory. Based on their contribution
they are allocated some fraction of the observatory time.

The time sums over given period and the fraction left for the observation is
adjusted accordingly. Suppose that we have two groups sharing time on the
telescope, both having equal share (50\%) of the telescope time. Then if two
nights are scheduled, and one group receives first night for its observations,
the other group shall get remaining full night.

Time allocated to the groups is accounted, and compared with the share values.
If some shared values drop bellow reasonable number, system must give higher
preference to this group in order to successfully fill requested share
fractions.

To formalise this mechanism consider $a$ accounts. Vector $A$ of length $a$
holds fraction of time allocated to each account. It is clear that

\[
\sum_{k=1}^{a}A[k] = 1
\]

Vector $OA$ of length $a$ holds seconds accounted for various groups. Total
time of all observations, $OT$, can then be calculated as 

\[
OT = \sum_{k=1}^{a}OA[k]
\]

If $OT > 0$, current percentage for a given account, $OC[k]$, is calculated as

\[
OC[k] = \frac{OA[k]}{OT}
\]

\section{Observation schedule}

Observation schedule is an ordered sets of targets. For each target, starting
time is provided. After target is selected for observation, telescope is slewed
to target position and observing sequence is executed. Shutter open time and
total observing time can be calculated for observing sequence provided with
target.

Schedule is a set of three vectors of length $s$. Vectors are $SS$, $ST$ and
$SL$. Vector $SS$ contains start time of observations. Vector $ST$ contains
targets which are scheduled for observation. And $SL$ contains observation loop
counts. For feasible schedule, following conditions must be fulfilled:

\[
\forall_{k=1}^{s - 1} SS[k+1] \geq SS[k] + T_s[ST[k]] + SL[k] * T_t[ST[k]]
\]

\[
\forall_{k=1}^{s}SL[k] \geq 1
\]

\[
SS[1] \geq N_s
\]

\[
SS[s] \leq N_e
\]

\section{Number of targets observed during night}

Observing schedule should try to visit as much targets during night as
possible. Schedule which contains only observations of two targets visited
through whole night is most probably not better then schedule with three, four
or more targets. That is because due to probability, more visited and observed
targets can bring more opportunities to discover new science and hence write a
good paper - and as mentioned at the beginning, the whole game is at the
ultimate end about publications.

On the other hand, an excess fragmentation of night time is weighted good.
Excess fragmentation will make long--duration observations highly improbable.
Long--duration observations are necessary for planet transits and other
science. The solution may be found in a careful examination of the possible
schedules and picking sometimes ones with fewer targets, but more
long--duration runs, and sometimes go for a large night fragmentation.

It must be also mentioned that big fragmentation naturally allows better time
distribution and hence creating a schedule which will fill accounted time of
various groups. So the system shall aim for a bigger fragmentation in order to
be able to better distribute remaining time.

There should be an objective night fragmentation, expressed in number of
targets visited during night. The better schedule is the one with number of
targets visited closer to this number.

\chapter{Time-dependent objective functions}

This section deals with problem of using various, usually time-dependent target
fitness functions to calculate fitness of the whole schedule.

As two different schedules can hold different number of targets, using sum of
observation fitness included in schedule will be useless. It is also important
to note that observation fitness can be different during duration of the
observation. For example value of observation position fitness calculated from
object altitude will change with a daily and other movement of the object on
the sky.

The first solution to those problems is to use average fitness calculated at
the midpoint of the observation duration, which can be expressed as:

\[
\frac{\displaystyle\sum_{k=1}^{s} f^{ST[k]} (SS[k] + T_s[ST[k]] + \frac{SL[k] * T_o[ST[k]]}{2})}{s}
\]

where $f^{T}(t)$ is value of the fitness function for target $T$ at time $t$,
and there are $s$ targets in the schedule. There are however still some
problems associated with this approach:

\begin{itemize}

\item as there are multiple fitness functions, it does not present single
objective, but rather multiple objectives

\item fitness value at the various times can different significantly from
fitness value at the observation middle time

\item the functions does not differentiate between schedules with higher number
of observations and those with fewer observations

\end{itemize}

To handle differences due to time used for calculating observing fitness,
minimum can be used. So the function then becomes:

\[
\frac{\displaystyle\sum_{k=1}^{s} min_{t \in < SS[k] + T_s[ST[k]], SS[k] + T_s[ST[k]] + SL[k] * T_o[ST[k]] >} (f^{ST[k]} (t))}{s}
\]

This function truly evaluates targets merit functions. Averaging will make sure
that schedules with fewer observations will not be disadvantaged against
schedules with more observations.

Most probably fitness functions shall be evaluated separately. As some
objectives are contradicting it is impossible to construct schedule with only
the best observations at the best times. There will be multiple paths to choose
from, and the whole play is about sufficient balance between different objectives.

\chapter{Multiobjective scheduling optimisation}

Multiobjective scheduling optimisation is discussed in great detail in
\cite{multiobjsch}. Here are discussed various method to select best schedule in
problems with multiple independent objective functions. The possible solutions
are reviewed bellow.

\section{Weighted single objective function}

This is probably the simplest approach. Objective functions are multiplied with
weighting factors and summed together to form a single objective function.
Scheduling algorithm then search for schedule with highest single objective
function.

The major disadvantage of this approach is necessity of finding correct weight
factors.

\section{Single objective function, move others objectives to constraints}

This approach picks the most significant objective as the single objective.
Other objectives are then used as schedule constraints.

Major disadvantage of this approach is in specifying correct constraints for
objective functions which are not used as a single objective. When the
constraint range is too narrow, there is a risk of loosing some good solution
because they will slightly not fit inside the range. If the range is too wide,
there is a risk of finding schedules way from the best one.

\section{Searching Pareto optimal solutions}

Pareto\cite{pareto} optimality is named after Vilfred Pareto. This method
search for all nondominant solutions. It overcomes disadvantages of both
previous approaches by finding subsurface in the solution space with the best
possible tradeoffs between various objective functions. It does not need any
weight factor nor correctly picked constraints.

Genetics algorithms are very good in finding Pareto optimal subsurface. As the
algorithm always operates with multiple solutions, they can represent multiple
points on Pareto optimal subsurface. So the genetic algorithm naturally fits in
Pareto search.

Following section describes one of the genetics algorithm variants, know as
Nondominated Sorting Genetic Algorithm II - \mbox{NSGA II}.

\chapter{Nondominated Sorting Genetic Algorithm II}

Scheduler uses Nondominated Sorting Genetic Algorithm II (NSGA-II) developed by
Deb et al. A short description provided by Deb et. al. \cite{Deb00} is the
following:

The step-by-step procedure shows that NSGA-II algorithm is simple and
straightforward. First, a combined population $R_t = P_t \cup Q_t$ is formed.
The population $R_t$ is of size $2N$. Then, the population $R_t$ is sorted
according to non-domination. Since all previous and current population members
are included in $R_t$, elitism is ensured. Now, solutions belonging to the
best non-dominated set $F_1$ are of best solutions in the combined population
and must be emphasised more than any other solution in the combined
population. If the size of $F_1$ is smaller then $N$, we definitely choose all
members of the set $F_1$ for the new population $P_{t+1}$. The remaining
members of the  population $P_{t+1}$ are chosen from subsequent non-dominated
fronts in the order of their ranking. Thus, solutions from the set $F_2$ are
chosen next, followed by solutions from the set $F_3$, and so on. This
procedure is continued until no more sets can be accommodated. Say that the
set $F_i$ is the last non-dominated set beyond which no other
set can be accommodated. In general, the count of solutions in all sets from
$F_1$ to $F_i$ would be larger than the population size. To choose exactly $N$
population members, we sort the solutions of the {\it last} front $F_i$ using
the crowded-comparison operator $\prec_n$ in descending order and choose the
best solutions needed to fill all population slots. The NSGA-II procedure is
also shown in Fig. \ref{pseudo}. The new population $P_{t+1}$ of size $N$ is
now used for selection, crossover, and mutation to create a new population
$Q_{t+1}$ of size $N$. It is important to note that we use a binary tournament
selection operator but the selection criterion is now based on the
crowded-comparison operator $\prec_n$. Since this operator requires both the
rank and crowded distance of each solution in the population, we calculate
these quantities while forming the population $P_{t+1}$, as shown in the above
algorithm.

The components of the \mbox{NSGA-II} scheduling structure are described as
follows:

\begin{figure}[htb]
\centering
\includegraphics[scale=0.35]{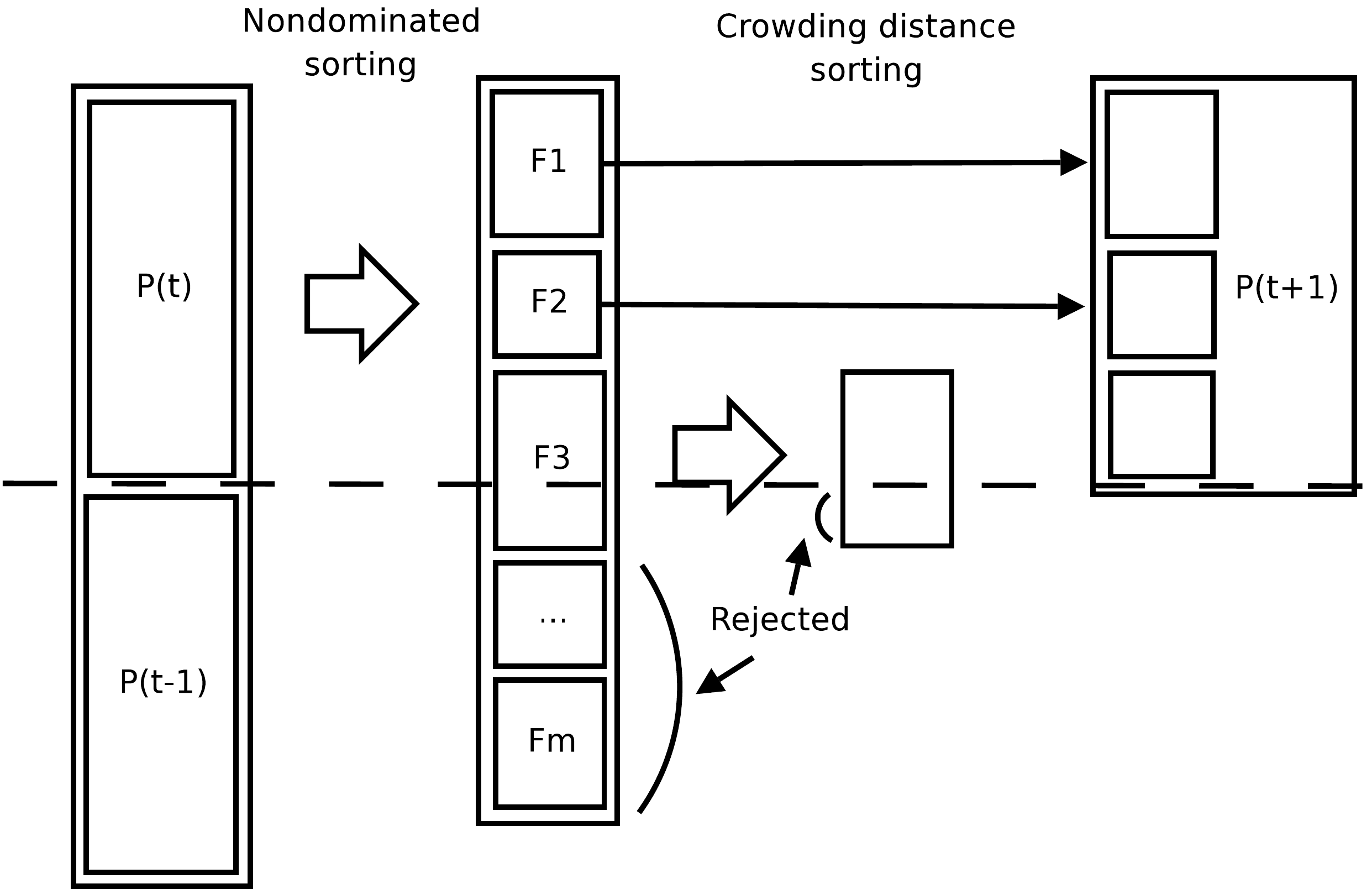}
\caption{The \mbox{NSGA-II} algorithm \cite{Deb00}.}
\label{pseudo}
\end{figure}

\section{Chromosome Representation}

\mbox{RTS2 NSGA-II} scheduling encodes only feasible schedules in each
chromosome. Chromosomes are implemented as an array of observing entries. The
gene in chromosome is a record containing starting date, duration and pointer
to ticket\footnote{which contains observation details - target, account etc.}.
The initial population consists of a set of random schedules, generated using
random number generator. The set of all nondominant chromosomes of the final
population represents an optimal schedules.

\section{Genetic Operators}

\mbox{RTS2 NSGA-II} scheduling applies crossover and mutation operators with a
given probability over the chromosomes composing the GP population. The
crossover operator consists of the following steps:

\begin{itemize}

\item pick a random time $T_{cross}$between night start and night end.

\item construct beginning of the resulting schedule by using observations from first schedule till $T_{cross}$

\item add schedules to the resulting schedule from second schedule, starting from $T_{cross}$.

\item repair resulting schedule, so it is feasible - adjust schedule starting
time, and schedules duration. If schedule duration cannot be adjusted, remove
shortest schedule.

\end{itemize}

The mutation operators used in \mbox{RTS2 NSGA-II} scheduling implementation are those:

\begin{itemize}

\item Delete a random selected observation

\item Change ticket entry of a random selected observation

\item Adjust duration by a random time of a randomly selected observation

\end{itemize}

After the mutation is performed, resulting schedule is repaired by same
algorithm used for repairs after crossover.

\section{Selection}

\mbox{RTS2 NSGA-II} scheduling employs a crowded binary tournament constraint
-- dominated selection operator \cite{Deb00}. Assuming that every individual in
the population has three attributes: number of violated constraints
($i_{violation}$, nondomination rank ($i_{rank}$) and crowding distance
($i_{distance}$), the crowded constraint -- dominated operator $\prec_{cc}$ is defined as

\[
i \prec_{cc} j \iff
\begin{cases}
(j_{violation} > 0 $ and $ i_{violation} < j_{violation}) \\
$or $  (i_{rank} < j_{rank})  \\
$or $  (i_{rank} = j_{rank} $ and $ i_{distance} > j_{distance})
\end{cases}
\]

See \cite{emocc} and \cite{Deb00} for a complete description.

\section{Constraints}

Selection of crowded constraint -- dominated selection operator allows easy
addition of new constraints. Constraint functions returns integer values, which
tell how many constraints are violated. The objective of algorithm is to
minimise this value, so there will be as few constraints as possible. The following 
constraints can be used:

\begin{itemize}

\item{\bf Visibility.} Observation violates visibility constraint, if it is not
visible during its scheduled time.

\item{\bf Schedule time.} Observing ticket might provide time during which
observation should be carried. If observation is not carried in the specified
time interval, it breaks schedule time constraint.

\item{\bf Unobserved tickets.} If time period during a ticket should be
observed intersect with interval being scheduled and it is not selected for
observation, it violates this constraint.

\item{\bf Number of observations per ticket.} Some tickets might provide number
of observation required to be performed of the target. Schedule violates this
consists if more observations of the ticket are schedule.

\end{itemize}

\section{Fitness functions}

One of the principal advantages of NSGA-II multi-objective algorithm is ability
to easy add new fitness functions. Following fitness functions are used: {\it
altitude, observation distance, account, target diversity, observation
diversity}. The diversity functions conflicts with observation distance - if
schedule has better diversity, it has worse observation distance and vice
versa. The implementation works to maximalize fitness functions and minimalize
constraints violations. In the fitness functions description is provided note
if target is to maximalize or minimalize its value. If not specified otherwise,
it is assumed that if objective is to minimalize fitness value, inverted
value is used in algorithm for maximalization.

\begin{itemize}

\item{\bf Altitude merit.} Altitude merit is calculated as ratio of mid
altitude to maximal possible altitude which target can have during night. For a
given observing ticket it is calculated as:

\[
f_h(h) =
\begin{cases}
0 & if h \leq TA_{min} \\
\frac{h - max (Anight_{min}, TA_{min})}{Anight_{max} - max (Anight_{min}, TA_{min})} & otherwise
\end{cases}
\]

For final schedule merit is used average of those ticket functions. Objective
is to maximalize this value.

\item{\bf Observation distance merit.} This merit is calculated as sum of
distance of the telescope travelled. Its  purpose is to minimalize time
telescope will spend moving from one location to the other. For a single
observation it is calculated as:

\[
f_d() = angularDistance (Position^{previous}_{end}, Position^{current}_{start})
\]

Sum of the individual values is used. Objective is to minimalize this value.

\item{\bf Account merit.} Account merit is calculated as ratio of observed
schedule account use versus requested account use:

\[
AD = \sum_{k=1}^{a} \frac{|OC[k] - OA[k]|}{OA[k]}
\]

where $AD$ is sum of proportional differences of requested and observed
accounting.

Objective of the scheduling algorithm is to find schedule with minimal deviation
from requested time share. Time share is accounted usually by longer
intervals, months, semesters or year. So the scheduling algorithm shall give
lower priority on fairness of the selection at the beginning of the accounting
period then at the end of the accounting period.

\item{\bf Target diversity merit.} Target diversity merit is calculated as
number of targets observed in the schedule. Objective is to maximalize this
function.

\item{\bf Observation diversity merit.} Observation diversity merit simply
counts observing entries in the schedule. Objective is to maximalize this
function.

\end{itemize}

\chapter{Implementation}

Because RTS2\cite{rts2} is mostly coded in \cite{GNUC++}, choice of the
language in which scheduler shall be written was pretty obvious.
Coding was done in the \href{http://www.vim.org}{Vim}\cite{vim} editor.
debugging was done using {\href{http://www.valgrind.org}{Valgrind}} and
\href{http://sourceware.org/gdb}{GDB: The GNU Project Debugger}.

Code was documented using \href{http://www.doxygen.org}{Doxygen}. The design
relies as much as possible on standard template library provided by
\href{http://gcc.gnu.org/onlinedocs/libstdc++}{GNU libstc++}. 
\href{http://libnova.sourceforge.net}{LibNova} was used for various
astronomical calculations.

The implementation benefits from object oriented approach. It provides classes
which holds list of schedules = {\it GA population} = {\bf Rts2SchedBag},
schedules = {\it chromosomes} = {\bf Rts2Schedule} and observation entries =
{\it genes} = Rts2SchedObs. Observation targets are subclasses of {\bf Target}
class, created by a standard {\bf createTarget} call. Rts2SchedBag provides
methods for GA algorithms. Rts2Schedule provides methods for chromosome
evaluation.

Interface for testing was written as subclass of the standard
{\bf Rts2AppDb} class. The interface provides few options, and prints out
results in simple space separated format.  The output can be feed directly to
\href{http://www.gnuplot.org}{GNUPlot} plotting program. It is expected that
scheduling classes will be integrated to RTS2 as a standard library.

During development
\href{http://en.wikipedia.org/wiki/Iterative_and_incremental_development}{iterative
life cycle} was used. Small parts of the system were developed, tested and
results checked. The following sections document progress of development.

Development was initially committed to
\href{http://rts-2.svn.sourceforge.net/viewsvn/rts-2/branches/REL_0_8_0}{REL\_0\_8\_0}
branch. After firsts successful tests of GA code, branch was merged to trunk.

\section{Simple test}

\begin{figure}[H]
\centering
\includegraphics[width=1\linewidth]{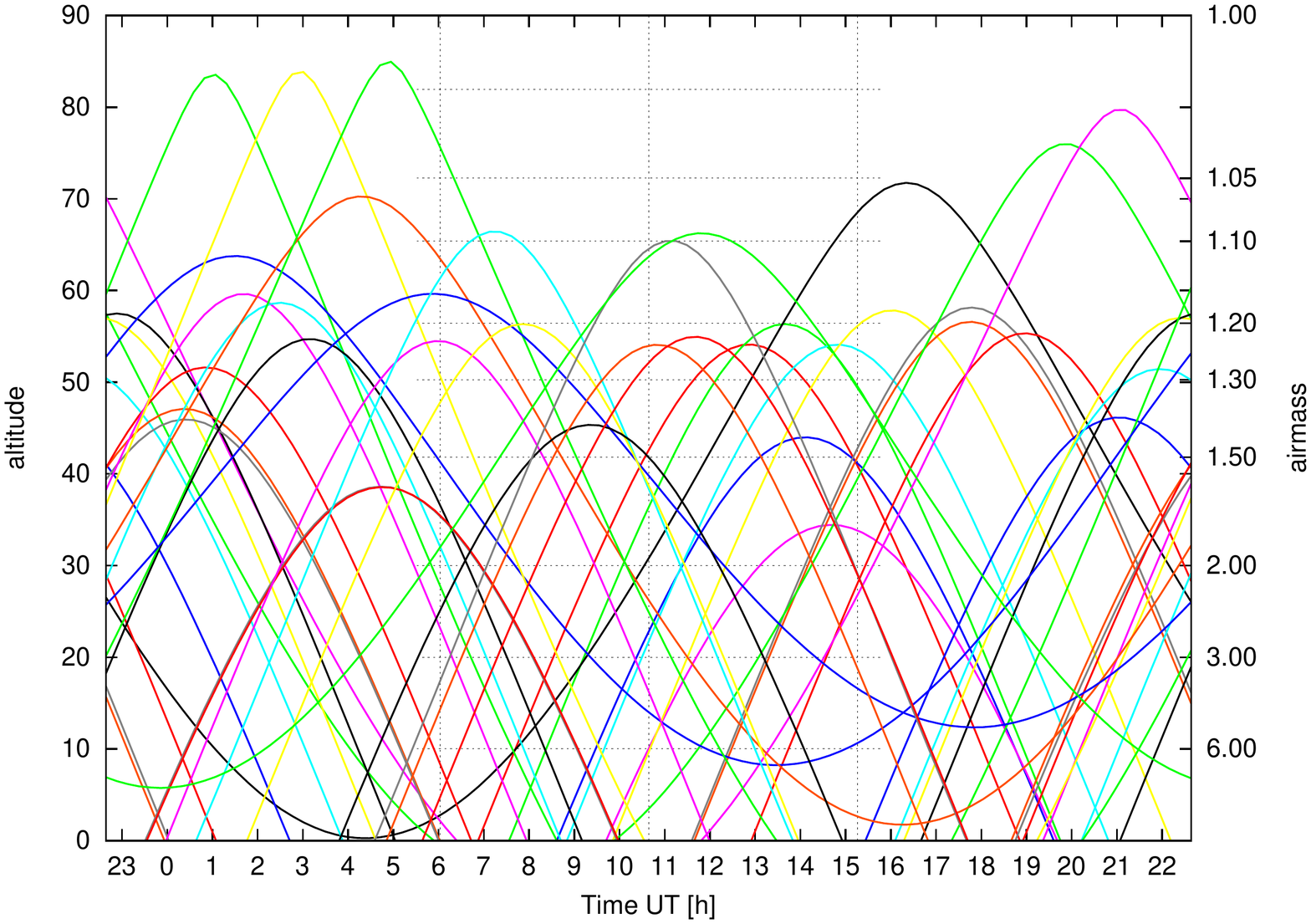}
\caption{Altitude of targets used for first tests as function of time.
\label{fig:ga_simple_alts}}
\end{figure}

First test was done on a simple genetics algorithm for selecting visible
schedules. Target set consists of flat field targets used to obtain calibration
observations. Plot of targets altitude as function of time observed from a site
at 36\degree north latitude are show in figure \ref{fig:ga_simple_alts}.  As
targets are distributed along the celestial equator, it is possible to observe
each target for 12 hours. 

Schedule consists of observations with predefined total time. Valid schedule
for this problem is any sequence of targets which fills requested time.

The algorithm work in the following steps:

\begin{itemize}

\item Initial random population is created
\item Elite population is chosen. Only the most fit schedules are drawn for
mating.
\item Mating is performed using roulette wheel selection. Crossing operator is
simple two fold crossing -- random number $r$ smaller then number of
observations in a schedule is drawn. Two child are created -- one with first
$r$ observations from the first parent and rest from second parent, the other
with opposite parents chromosomes used.
\item Mutation is performed. Random observation entry is picked and replaced by
another random observation entry.
\item Population number is increased
\item If population number is bellow predefined population maxima, go to step
2.
\end{itemize}

\begin{figure}[ht]
\centering
\includegraphics[width=1\linewidth]{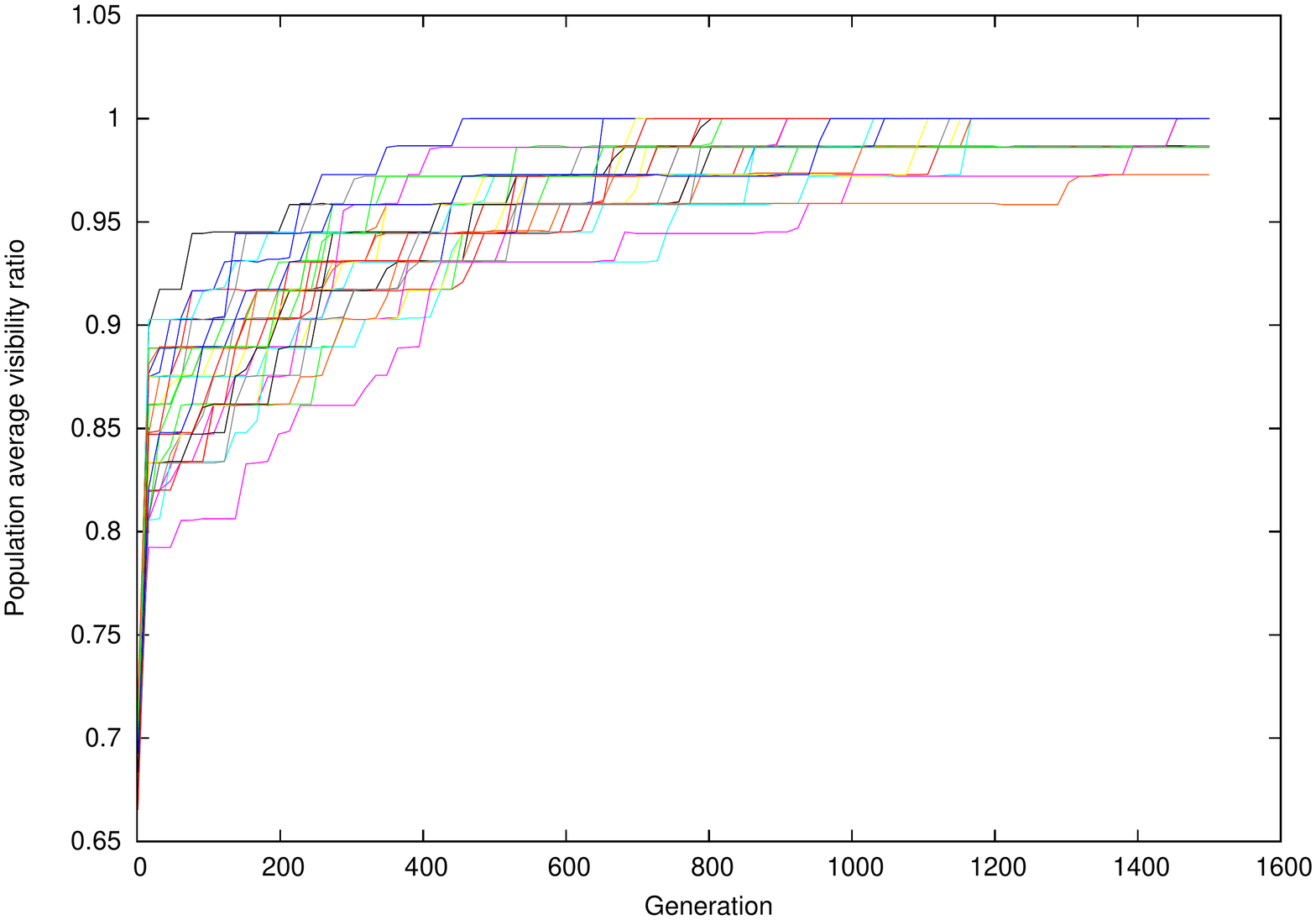}
\caption{Convergence of the population average visibility ratio.
\label{fig:ga_simple}}
\end{figure}

\begin{figure}[ht]
\centering
\includegraphics[width=1\linewidth]{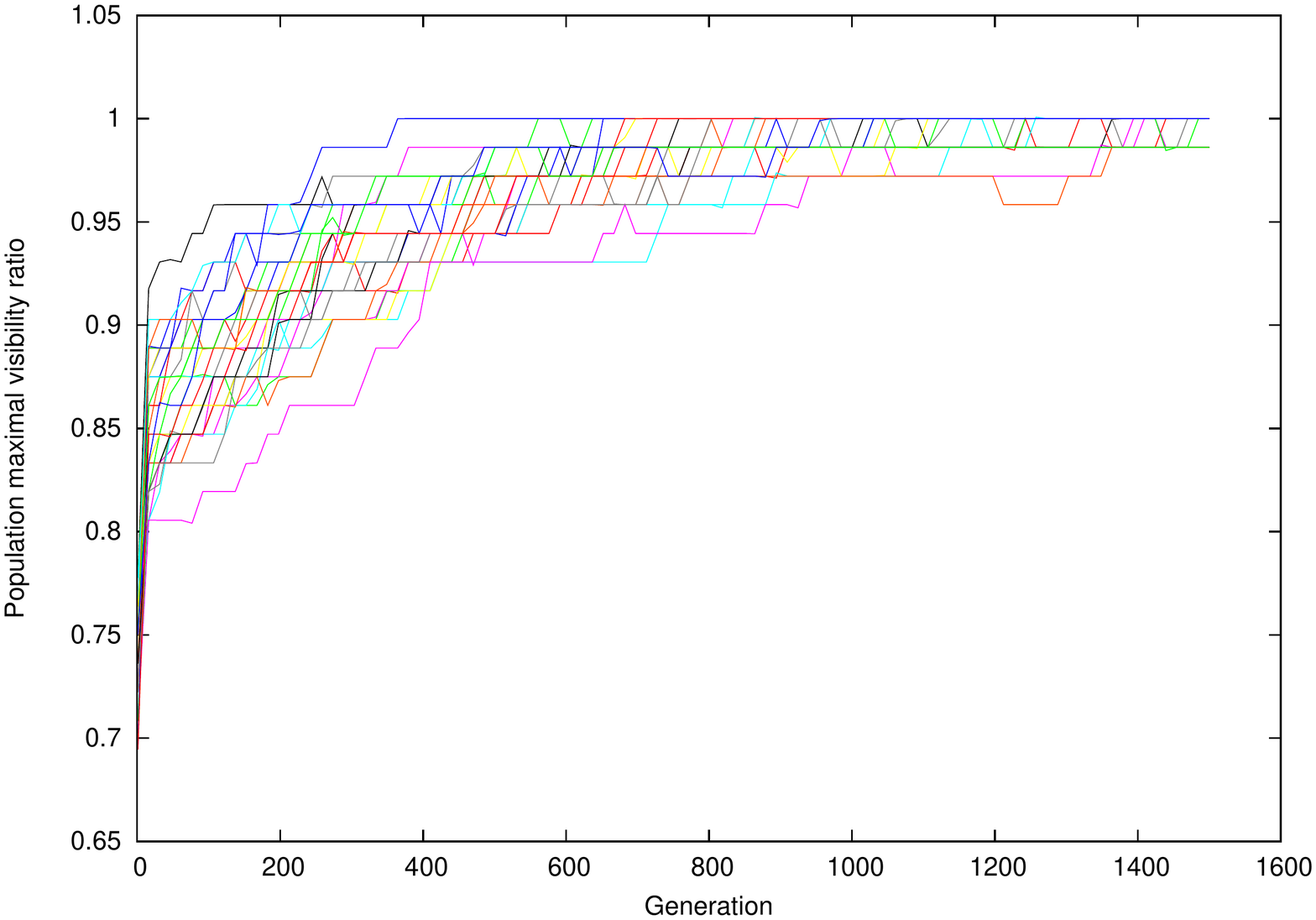}
\caption{Convergence of the maximal visibility ratio.
\label{fig:ga_simple_max}}
\end{figure}

Only schedule visibility ratio was used as fitness criteria. The results
confirmed correctness of genetic algorithm implementation. Results for 30 test
runs are presented in figures \ref{fig:ga_simple} and \ref{fig:ga_simple_max}. It can be clearly seen that:

\begin{itemize}

\item it works - the visibility fitness converge towards 1

\item the population converges pretty fast

\item as targets are distributed along celestial equator, the average visibility fitness of a random observation schedule is 0.5.

\end{itemize}

These tests confirms quick convergence of genetic algorithm. Quite nice results
was a quick convergence of entire population to global maxima. Those results
provided firm ground for further improvements.

\chapter{Results}

\begin{figure}[hp]
\includegraphics[width=1.25\linewidth]{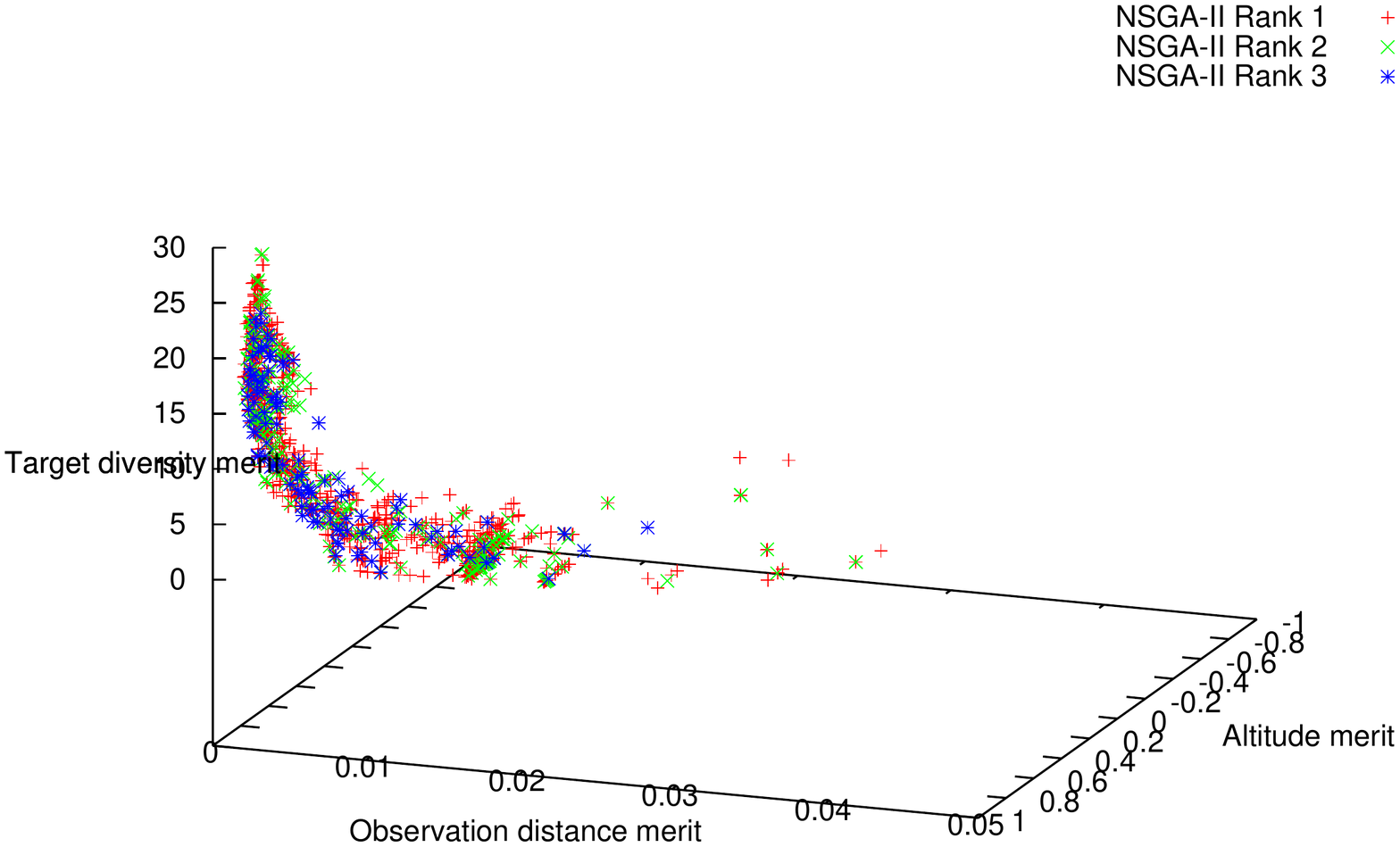}
\centering
\caption{Pareto front, population = 1000, generations = 100}
\label{fig:1000_100}
\end{figure}

\begin{figure}[hp]
\includegraphics[width=1.25\linewidth]{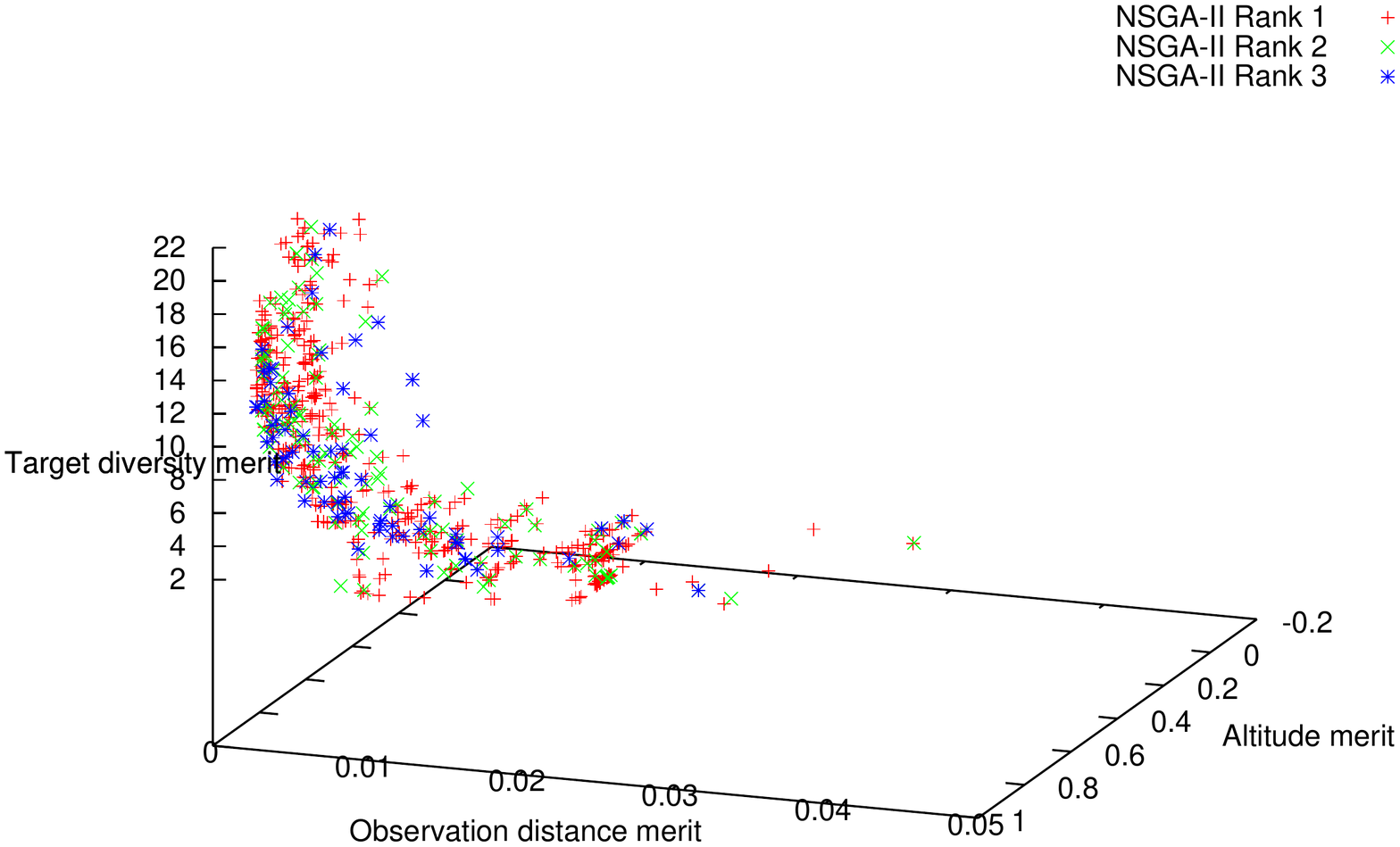}
\centering
\caption{Pareto front, population = 500, generations = 100}
\label{fig:500_100}
\end{figure}

\begin{figure}[hp]
\includegraphics[width=1.25\linewidth]{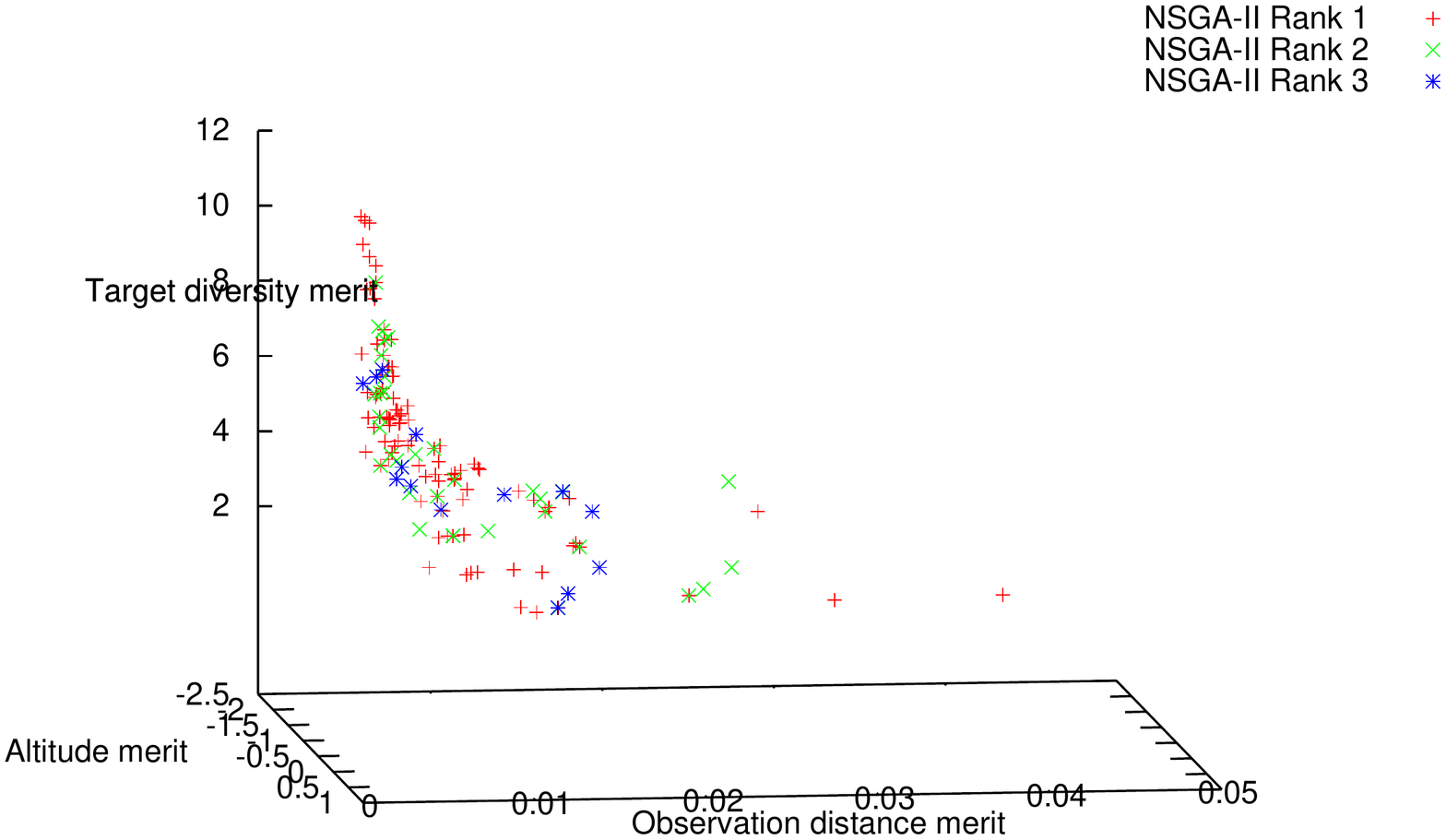}
\centering
\caption{Pareto front, population = 100, generations = 100}
\label{fig:100_100}
\end{figure}

The algorithm clearly identify Pareto optimal fronts. Figures
\ref{fig:1000_100}, \ref{fig:500_100} and \ref {fig:100_100} shows altitude,
observation distance and target diversity merits for different populations
sizes.

\begin{figure}[ht]
\includegraphics[width=1\linewidth]{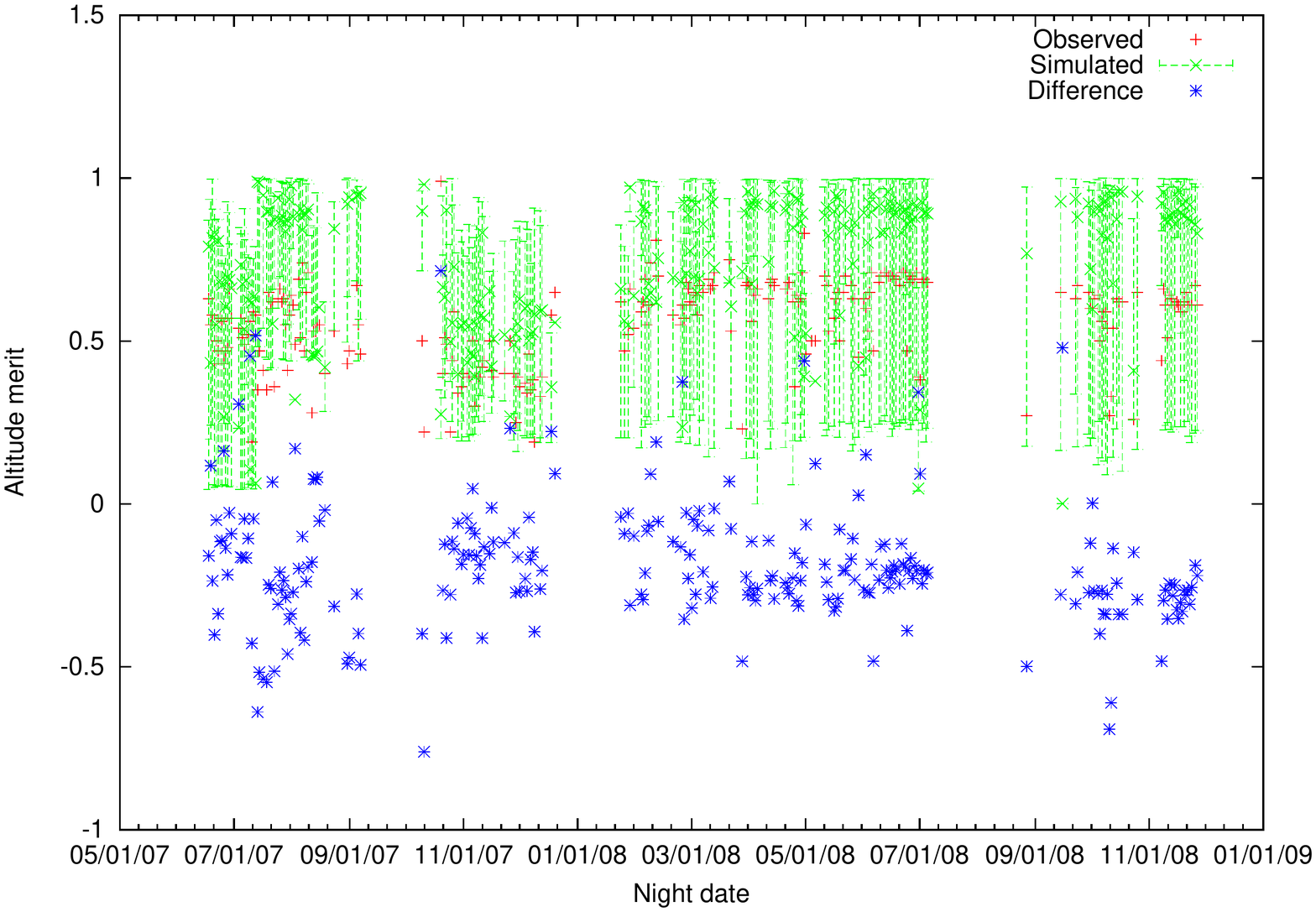}
\centering
\caption{Altitude merits of a used single merit algorithm versus new \mbox{NSGA-II} scheduling}
\label{fig:bootes1b_alt}
\end{figure}

\begin{figure}[ht]
\includegraphics[width=1\linewidth]{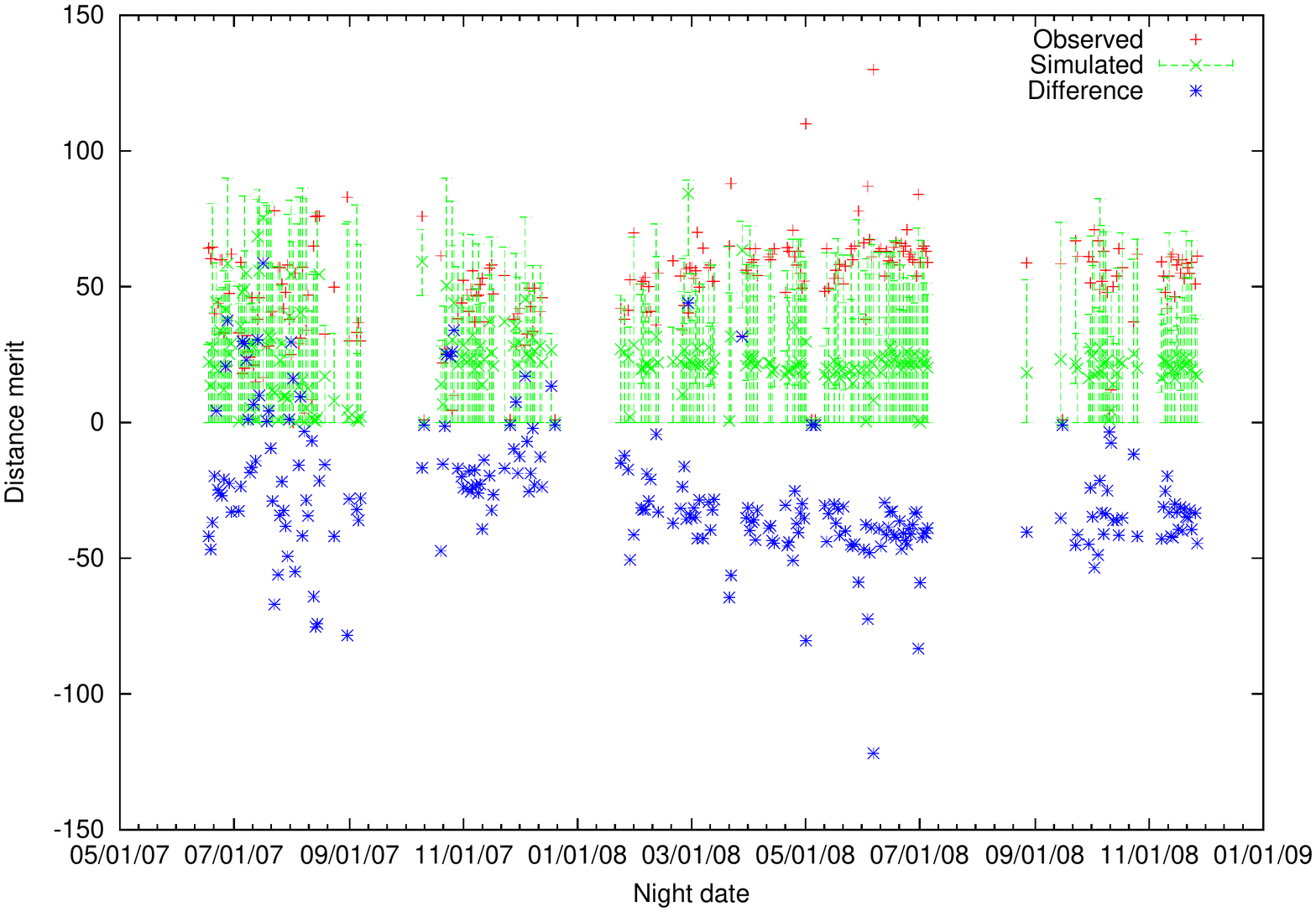}
\centering
\caption{Distance merits of a used single merit algorithm versus new \mbox{NSGA-II} scheduling}
\label{fig:bootes1b_dist}
\end{figure}

Tests against currently used single objective algorithm were carried. Figures
\ref{fig:bootes1b_alt} and \ref{fig:bootes1b_dist} shows simulated altitude
merit and distance merit functions versus observed merit and distance functions.
Unfortunately account merit cannot be calculated, as this merit was recently
introduced.

Altitude merit is in proportional units, where 1 means optimal altitude of the
observations. Distance merit is in degrees - there lower value means lower slew
times, and so better schedule. In both graphs, blue dots are difference between
observed schedule and average of Pareto front schedules. There lower difference
value means better \mbox{NSGA-II} scheduling simulation then current scheduler.

\chapter{Conclusion}

In this work was presented a novel approach to telescope scheduling, using
Pareto optimal search genetic algorithm. Having Pareto optimum provides
experienced observers with overview which observations are possible.

Complex autonomouse telescope scheduling is a difficult task. It requires
continuous adjustment of objectives, so the observatory remains productive for
a various science goals. Also new observatory constraints can be introduced.
Presented approach provides an easy and robust way how to add new objectives and
constraints without a need to invest time and effort towards discovering
heuristics and rules which will make scheduling working better.

Scheduling network of the autonomouse observatories is a magnitude more
difficult then scheduling of a single observatory. Yet the approach outlined in
this work looks promising and provides solid base for a development of an
algorithm for network scheduling.

The software is ready for live use on the telescopes of \mbox{RTS2} network. It
is expected that it will be used in production during first quarter of 2009.

\chapter{Further work}

This work presents solid foundations for observatory and network scheduling.
The expected further work is related to further development of the \mbox{RTS2}.
This includes development of the central planning and monitoring facility,
which will enable observers to continuously monitor network performance. This
will also solve various operational issues and enables network scheduling. It
is expected that the network scheduling functionality will be added to network
in second quarter of 2009, at the time when \mbox{Bootes 3} telescope, located
on New Zealand, will start routine operations.

\backmatter

\bibliography{scheduling}
\bibliographystyle{plain}

\end{document}